\documentclass[conference]{IEEEtran}
\IEEEoverridecommandlockouts
\usepackage{cite}
\usepackage{amsmath,amssymb,amsfonts}
\usepackage{algorithmic}
\usepackage{graphicx}
\usepackage{textcomp}
\usepackage{xcolor}
\usepackage{float}
\usepackage{booktabs}
\usepackage{acronym}
\newacro{CNN}[CNN]{Convolutional Neural Network}
\newacro{BoF}[BoF]{Bag of Features}

\def\BibTeX{{\rm B\kern-.05em{\sc i\kern-.025em b}\kern-.08em
    T\kern-.1667em\lower.7ex\hbox{E}\kern-.125emX}}
\begin{document}

\title{Convolutional autoencoder-based \\multimodal one-class classification}
\author{\IEEEauthorblockN{Firas Laakom\IEEEauthorrefmark{1},
Fahad Sohrab\IEEEauthorrefmark{1},
Jenni Raitoharju\IEEEauthorrefmark{4} \IEEEauthorrefmark{2},
Alexandros Iosifidis\IEEEauthorrefmark{3} and
Moncef Gabbouj\IEEEauthorrefmark{1}}
\IEEEauthorblockA{\IEEEauthorrefmark{1}Faculty of Information Technology and Communication Sciences, Tampere University, Finland}
\IEEEauthorblockA{\IEEEauthorrefmark{4}Quality of Information, Finnish Environment Institute, Finland }
\IEEEauthorblockA{\IEEEauthorrefmark{2}Faculty of Information Technology,  University of Jyväskylä, Finland }
\IEEEauthorblockA{\IEEEauthorrefmark{3}Department of Electrical and Computer Engineering, Aarhus University, Denmark }

}
\maketitle
\begin{abstract}
One-class classification refers to approaches of learning using data from a single class only. In this paper, we propose a deep learning one-class classification method suitable for multimodal data, which relies on two convolutional autoencoders jointly trained to reconstruct the positive input data while obtaining the data representations in the latent space as compact as possible. During inference, the distance of the latent representation of an input to the origin can be used as an anomaly score. Experimental results using a multimodal macroinvertebrate image classification dataset show that the proposed multimodal method yields better results as compared to the unimodal approach. Furthermore, study the effect of different input image sizes, and we investigate how recently proposed feature diversity regularizers affect the performance of our approach. We show that such regularizers improve performance.
\end{abstract}
\section{Introduction}
\label{sec:intro}
Humans perceive the world using multiple senses, e.g., eyes to look and ears to listen. Such multimodal data provide information from different aspects. Inspired by the multi-sensory information integration ability of humans \cite{gazzaniga2006cognitive}, multimodal data collected from different sensors have been used in various deep learning methods \cite{de2022graph,wang2022improving,fu2022cma,mikriukov2022unsupervised,khokhlova2020cross}. Multimodal approaches have shown superior performance compared to unimodal approaches in various tasks \cite{huang2021makes}, e.g., object recognition \cite{wang2022improving}, audio-visual recognition \cite{fu2022cma,mikriukov2022unsupervised}, and image retrieval \cite{khokhlova2020cross}. In this work, we propose a multimodal approach for the one-class classification (OCC) problem \cite{khan2014one,ruff2018deep}.

OCC refers to the task of learning a binary classifier using data from a single class only \cite{tax2004support,sohrab2018subspace,sohrab2020ellipsoidal}. The decision function is inferred by utilizing data only from a particular class of interest, known as the positive or target class. Data belonging to any other class are not used during the training process and are considered to form an outlier or negative class \cite{tax2004support,tax2002one}. The aim is to construct a model that can accurately distinguish between unseen data belonging to the positive class or the negative class during inference. One-class classification techniques are useful in cases where data from the negative class is either difficult to collect or too diverse in nature that the accurate representative class cannot be formed \cite{sohrab2018subspace}. Numerous unimodal one-class classification methods have been proposed in the literature recently \cite{frikha2021few,sohrab2023graph,sohrab2020boosting,li2021anomaly}.

Although multimodal learning and one-class classification have been separately well studied, their intersection still needs to be further explored \cite {hu2021one,sohrab2021multimodal}. Our work addresses the multimodal OCC problem and presents a deep learning approach based on two convolutional autoencoders jointly trained to reconstruct the positive training data while keeping the representation in the feature space defined at the bottleneck of the architecture as compact as possible. Given that the positive data is mapped close to the origin in that feature space, during inference, we assume that any test point represented in the same feature space and falls close to the origin belongs to the positive class, and any data far from the origin is an outlier. In particular, we characterize the positive class region in the bottleneck representation of the autoencoder architecture as a hyper-sphere determined using a threshold based on the $L_2$ distance. We use the distance of the representation of a test data point to the center of this hyper-sphere as an anomaly score. We evaluate our approach using a subset of the multimodal macroinvertebrate image dataset \cite{raitoharju2018benchmark}. Experimental results show that our approach outperforms the unimodal approach. The main contributions of the paper are as follows: 

\begin{figure*}[t]
    \centering
    \includegraphics[width=0.98\linewidth]{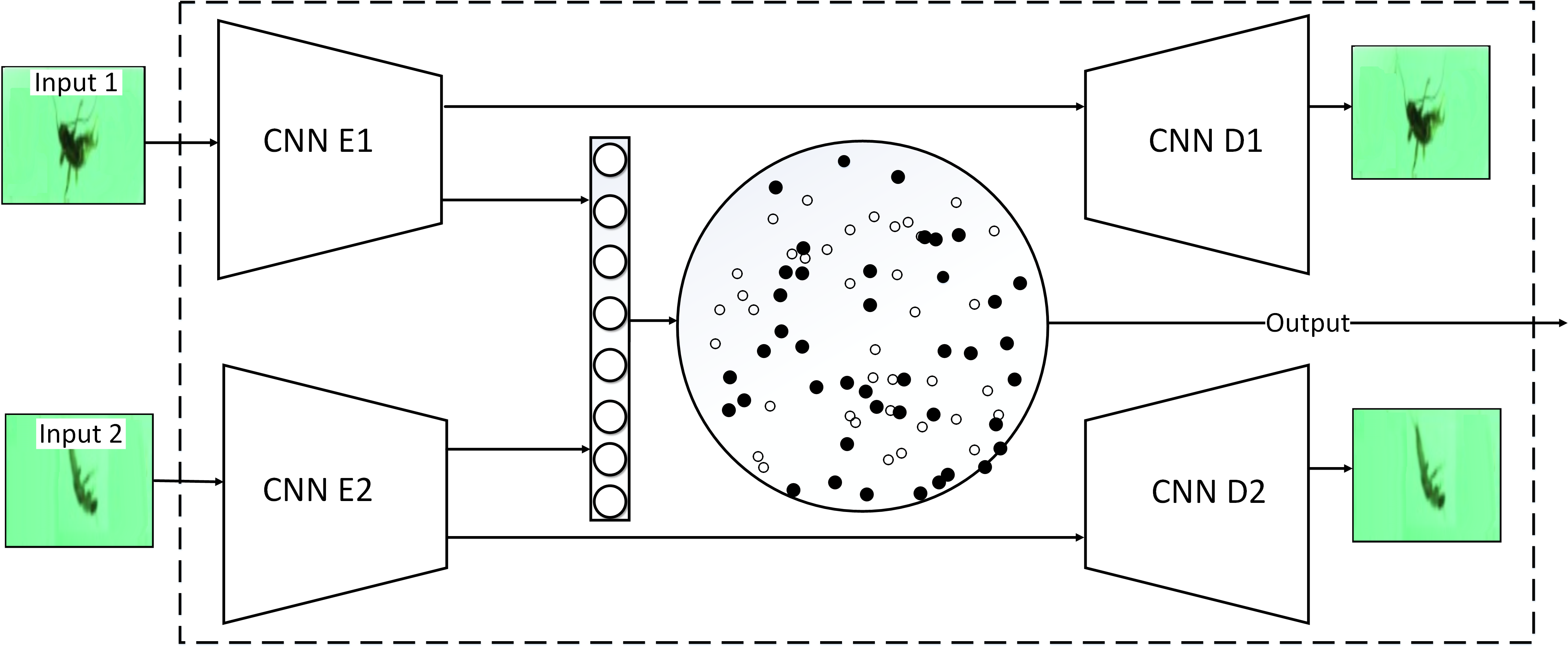}
    \caption{An overview of the proposed multimodal one-class classification approach. The aim is to keep the representation in the latent space as compact as possible while learning to reconstruct the inputs.  The input images from both modalities pass through the encoders CNN E1/E2; the outputs are concatenated to obtain the main embedding and are passed as input to the decoders CNN D1/D2.}
    \label{fig:68lndmrks}
\end{figure*}

\begin{itemize}
    \item We propose a multimodal one-class classification method using two jointly-trained convolutional autoencoders, which learn to map representations of two data modalities to a compact region at the feature space defined by the bottleneck of the network architecture.
    \item We evaluate our approach on the multimodal macroinvertebrate dataset \cite{raitoharju2018benchmark} and validate that multimodal learning yields better performance compared to the unimodal approaches. 
    \item We investigate how the input size affects the performance of the method and experiment with different feature diversity regularizers. We show that they can indeed improve the performance for this task.
\end{itemize}

\section{Proposed approach} \label{Proposed}
In this work, we consider the problem of one-class classification in the presence of multimodal data. We propose an approach that requires multimodal data only from the positive class in the training phase. Here, we describe the formulation for two modalities, while an extension to more modalities could be easily obtained. During training, the main target of our approach is to learn a compact mutual embedding of both modalities. Figure \ref{fig:68lndmrks} presents an overview of the method. Let $\{(x_i,x'_i)\}_{i=1}^N$ be the available training data from the positive class, where $x_i$ and $x'_i$ are the first and the second modality of the $i^{th}$ sample, respectively. Our model is composed of two autoencoders, one for each modality. Let $E_1(x_i) \in \mathbb{R}^{m_1 \times p_1 \times d_1 }$ be the convolutional output of the encoder on the first modality $x_i$ and $E_2(x'_i)\in \mathbb{R}^{m_2 \times p_2 \times d_2}$ be the encoder output of for the second modality. Based on these two outputs, we construct the joint latent representation $\phi_i$ of the sample $(x_i,x'_i)$ as follows:
\begin{equation} \label{output}
\phi_i=\phi(x_i,x'_i) = \text{concat}(\text{Flat}(E_1(x_i),\text{Flat}(E_2(x'_i))),
\end{equation}
where $\text{Flat}(\cdot)$ is the flattening operation, i.e, it flattens $E_1(x_i) \in \mathbb{R}^{m_1 \times p_1 \times d_1}$ into a $m_1p_1d_1$-dimensional vector. $\text{concat}$ is the vector concatenation operation compiling $\phi(x_i,x'_i) \in  \mathbb{R}^{(m_1p_1d_1 + m_2p_2d_2)} $ as the final representation of the input. 

Our main aim is to learn to map the input data from the positive class into a compact space. This is an objective commonly used by regression-based OCC models, and it is usually expressed by minimizing the distance of the latent representations to a pre-defined point \cite{leng2015one,iosifidis2017one,dai2019multilayer}. Using as the target point the origin, the objective becomes to minimize their $L_2$-norm:
\begin{equation} \label{eq1}
\small
   \frac{1}{N} \sum_{i=1}^N ||\phi_i||^2 = \frac{1}{N}   \sum_{i=1}^N || \text{concat}(\text{Flat}(E_1(x_i),\text{Flat}(E_2(x'_i))||^2.
\end{equation}
By minimizing \eqref{eq1}, the model learns to map the samples from the positive class into a hyper-sphere centered at the origin. In the test phase, any sample that falls close to the origin is assigned to the positive class, and the rest are classified as anomalies. 
\begin{table*}[h!] 
\caption{The Recall, the Precision at rank n (P@n), and the Area Under the Curve (ROC) of the three different models tested on the four one-class tasks. We also report the average performance over the four tasks for each model.}\label{tab:multimodal_results}
\centering
\begin{tabular}{c c c c c c c  c c c } 
Normal Class  &  \multicolumn{3}{c}{ unimodal (left)}  & \multicolumn{3}{c}{unimodal (right)}  & \multicolumn{3}{c}{ours (multimodal)} \\
 \hline\hline
 & Recall & P@n & ROC  & Recall & P@n & ROC  & Recall & P@n & ROC  \\ 
 \cmidrule(r){2-4}  \cmidrule(r){5-7}   \cmidrule(r){8-10}
 \textit{Leptophlebia sp.} & 0.963 & 0.921 & 0.806 & 0.888 & 0.897& 0.535 & 0.938 & 0.921  &0.742  \\
 \textit{Baetis rhodani} & 0.961 & 0.893 & 0.428 & 0.882 & 0.892&  0.330  & 0.961 & 0.907& 0.582 \\
\textit{ Elmis aenea larva} & 0.844 & 0.900 & 0.367 & 0.896 & 0.900 & 0.456 &  0.935 & 0.899 & 0.427    \\
\textit{Oulimnius tuberculatus larva} & 0.896 & 0.897 & 0.444 & 0.909& 0.919 & 0.646 & 0.896 & 0.897 & 0.560    \\
 \hline
 Average  & 0.916  &0.903 & 0.511 & 0.894 & 0.902 & 0.49 & \textbf{0.932} & \textbf{0.906} & \textbf{0.578} \\
  \hline
\end{tabular}
\end{table*}
Minimizing \eqref{eq1} can lead to a degenerate solution, i.e., the model learns to map all inputs to the origin and thus fails to distinguish between positive and anomalous samples. To avoid obtaining such solutions, we propose to augment our model using two decoders (one for each modality), aiming to learn to reconstruct the inputs. As illustrated in the right part of Figure \ref{fig:68lndmrks}, the outputs of the encoders $E_1(x_i)$ and $E_2(x'_i)$ are passed through to the decoders $D_1$ and $D_2$. To incorporate the reconstruction objective into the training, we propose to augment the loss in \eqref{eq1} using the mean squared loss. The final loss used to train the network can be expressed as follows: 
\begin{multline} \label{eq2}
   L :=  \frac{1}{N} \sum_{i=1}^N \Big( ||\phi_i||^2  +  ( || D_1(E_1(x_i)) - x_i ||^2 +  \\
     || D_2(E_2(x'_i)) - x'_i ||^2 )\Big).
\end{multline}
The weights of $E_1$ and $E_2$ are shared as well as the weights for $D_1$ and $D_2$, and they can be trained in an end-to-end manner using gradient-based optimization by minimizing \eqref{eq2}. The first term of the loss forces the model to learn a compact representation for both modalities of the same sample in the bottleneck of the architecture, while the second and third terms regularize the model to avoid degenerate and undesired solutions. 

In the test phase, we discard the decoder part. Given a test sample $(y,y')$, we compile its feature output $\phi$, as expressed in \eqref{output}. The distance of the latent representation of the data point from the origin can be used as an anomaly score. Based on this distance, we can assign the sample to the corresponding class (positive class or negative class). To this end, we need to determine a threshold $\tau$, which is used to define the hyper-sphere enclosing the positive class. The value of $\tau$ can be obtained using the training data. As all the training data is from the positive class, $\tau$ can be set to $95^{th}$ percentile of the feature norms of the training data $\{ \phi_i\}_{i=1}^N$. Then, given the test sample $(y,y')$, if $\phi(y,y')\leq \tau$, it is considered to be from the target class. Otherwise, it is considered an anomaly.  

It should also be noted that, although in this work, we use shared weights for $E_1$ and $E_2$, it is possible to use a different model for every modality. However, one-class classification tasks usually have scarce data \cite{seliya2021literature,sohrab2018subspace}. Using shared weights reduces the total number of parameters and acts as a regularization, which makes our model suitable for learning from a limited amount of data. 
\section{Experiments and Discussion}
In this section, we evaluate the performance of the proposed multimodal one-class classification method. To validate that multimodal learning helps in the context of one-class classification, we also report the performance of the corresponding unimodal one-class classification model.
\subsection{Dataset}
We used a subset of the multimodal image classification dataset of benthic macroinvertebrates, FIN-Benthic \cite{raitoharju2018benchmark}. In particular, we used data from 4 classes as presented in Table \ref{tab:dataset}. Each sample point is presented with two RGB images (which act as the two modalities) from two perpendicular viewpoints. Using this dataset, we constructed four different one-class classification tasks. In each task, data from a single class (out of the four) is considered the normal class, and the remaining three classes are combined to form the anomaly class. In each of the four experiments, we used $66\%$ of normal class data as training data, and we held the rest along with the anomaly data (the remaining three classes) as our test data. All the images were resized to $32\times32$ pixels.
\begin{table}[h] 
\caption{Number of samples for each of the four tasks used in our experiments.}\label{tab:dataset}
\centering
\begin{tabular}{||c c||} 
 \hline
 Class name & \# Samples \\  
 \hline\hline
\textit{Leptophlebia sp.} & 240   \\ 
 \hline
 \textit{Baetis rhodani} & 230   \\ 
 \hline
\textit{ Elmis aenea larva} & 231   \\ 
 \hline
\textit{Oulimnius tuberculatus larva} & 231  \\ 
 \hline
\end{tabular}
\end{table}

\subsection{Experimental protocol and results}
Our implementation is based on \cite{zhao2019pyod}. To train our models, we use Adam optimizer with a learning rate of 0.001 and weight decay of $10^{-3}$. The training is conducted with 4 epochs and a batch size of 32. The input image size is $32\times 32$ pixels. For $E_1$ and $E_2$ in Figure \ref{fig:68lndmrks}, we used a fully convolutional model which consists of three blocks of convolution, batch normalization, \textit{maxpooling}, and \textit{dropout} layers. All the convolution filters have a size of $3\times 3$ and were selected to be 64, 32, and 16 in the first, second, and third layers, respectively. For the decoder part, i.e., $D_1$ and $D_2$  in Figure \ref{fig:68lndmrks}, we used the corresponding symmetric layers.

To test the hypothesis that multimodal learning helps in the context of one-class classification, we also experimented with the unimodal variant of the method, i.e., using only the upper branch of the model in Figure \ref{fig:68lndmrks} and using images from one modality. This yields two competing methods, namely unimodal (left) and  unimodal (right), for the left and right modalities, respectively. 

In Table \ref{tab:multimodal_results}, we report the results for the multimodal model along with the two unimodal models on the four one-class classification tasks. We also report the average results over the four tasks. For each method, we report the Recall scores, the Precision at rank n (P@n), and the Area Under the Curve (ROC) \cite{zhao2019pyod}. 

As shown in Table \ref{tab:multimodal_results}, multimodal learning, indeed, yields better performance compared to both unimodal cases in all three metrics. For instance, in the average performance, the proposed multimodal model yields $0.067$ and $0.088$ improvement in ROC compared to the unimodal models using the left right images, respectively. We also note that on the four tasks, the worst ROC achieved by unimodal (left) and unimodal (right) models are $0.367$ and $0.330$, respectively, whereas, for the multimodal model, the lowest ROC corresponds to \textit{Elmis aenea larva} and is equal to $0.427$. 

\subsection{Ablation study}
\subsubsection{Input size}
Here, we investigate how the input size affects the performance of the proposed method. In particular, we experimented with the following input sizes: $32\times 32$, $64\times 64$, and $128\times 128$ pixels. The average results over the 4 tasks are reported in Table \ref{tab:input_size}. The best performance is achieved using $32\times 32$-pixel input size. We note that using a higher size improves the resolution. However, it adds more details and background noise to the images, which can render the learning task harder, especially with limited training data ($\sim$252 samples in our experiments). 

\begin{table}[H] 
\caption{Performance of the multimodal model using different input sizes.}\label{tab:input_size}
\centering
\begin{tabular}{|c c c c||} 
 \hline
Input size & Rec & P@n & ROC   \\ 
 \hline
 $32\times 32$ & 0.932 & 0.906 & 0.578 \\
 $64\times 64$ & 0.884 & 0.673& 0.512 \\
 $256\times 256$ & 0.864& 0.903 &  0.513 \\
 \hline
\end{tabular}
\end{table}

\subsubsection{Diversity regularizers}
Recently, there has been an interest in applying feature diversity regularizers to deep neural networks \cite{laakom2023wld,laakom2022reducing}, as they are proven to boost performance \cite{laakom2023learning}. Here, we evaluate how these regularizers affect the performance of the proposed method. In particular, we experimented with the three regularizers proposed in \cite{laakom2023wld}. The average results over the 4 tasks are reported in Table \ref{tab:div_results}. By comparing the results with the ones obtained without the regularizer, we note that applying diversity regularizers can indeed improve performance.

 \begin{table}[h]
\caption{Performance of the multimodal model trained with different feature diversity regularizers.}\label{tab:div_results}
\centering
\begin{tabular}{|c c c c||} 
 \hline
Approach & Rec & P@n & ROC   \\ 
 \hline
WLD-Reg (direct) & 0.939 & 0.906 & 0.579 \\
WLD-Reg (det) & 0.939 & 0.906& 0.580 \\
WLD-Reg (logdet) & 0.936& 0.907& 0.578  \\
 \hline
\end{tabular}
\end{table}
\section{Conclusion}
In this paper, we showed that multimodal learning helps in the one-class classification task. In particular, we proposed a deep learning method based on an autoencoder suitable for multimodal one-class classification. We proposed to learn to reconstruct the positive input training data while keeping the representation in the latent space defined at the bottleneck of the autoencoder architecture as compact as possible. In the test phase, the $L_2$-norm of the latent data representations can be used as an anomaly score. Experimental results using the multimodal macroinvertebrate image classification data show that the proposed multimodal approach indeed yields better results compared to the unimodal approaches. Moreover, in ablation studies, we investigated how the input size affects performance, and we experimented with feature diversity regularizers. We showed that such regularizers can improve the performance of the proposed method.
\section*{ACKNOWLEDGEMENT}
This work was supported by the NSF-Business Finland Center for
Visual and Decision Informatics (CVDI) project AMALIA.
Foundation for Economic Education (Grant number: 220363) funded the work of Fahad Sohrab at Haltian. The work of Jenni Raitoharju was funded by the Academy of Finland project TIMED (project 333497).

\end{document}